\documentclass[11pt,a4paper]{article}

\usepackage[hyperref]{acl2019}
\usepackage{times}
\usepackage{url}
\usepackage{amsfonts,amssymb}
\usepackage{amsmath}
\usepackage[utf8]{inputenc}
\usepackage[english]{babel}
\usepackage{graphicx}
\usepackage{xcolor}
\usepackage{tabularx}
\usepackage{multirow}
\usepackage{booktabs}
\usepackage{dsfont}
\usepackage{bm}
\usepackage{enumitem}
\usepackage{amsthm}
\usepackage{array}
\usepackage{anyfontsize}

\aclfinalcopy 
\def\aclpaperid{962} 

\newtheorem{thm}{\protect\theoremname}
\newtheorem{lma}{\protect\lemmaname}

\providecommand{\theoremname}{Theorem}
\providecommand{\lemmaname}{Lemma}

\title{Relation Embedding with Dihedral Group in  Knowledge Graph}

\author{Canran Xu \thanks{\protect\phantom{\footnotesize 1}Equal contribution.} \\
  eBay Inc. \\
  \tt canxu@ebay.com
\And Ruijiang Li \footnotemark[1] \\
  eBay Inc. \\
  \tt ruijili@ebay.com
\\}

\date{}

\begin{document}
\maketitle
\begin{abstract}
Link prediction is critical for the application of incomplete knowledge graph (KG) in the downstream tasks. As a family of effective approaches for link predictions, embedding methods try to learn low-rank representations for both entities and relations such that the bilinear form defined therein is a well-behaved scoring function. Despite of their successful performances, existing bilinear forms overlook the modeling of relation compositions, resulting in lacks of interpretability for reasoning on KG. To fulfill this gap, we propose a new model called \textit{DihEdral}, named after dihedral symmetry group. This new model learns knowledge graph embeddings that can capture relation compositions by nature. Furthermore, our approach models the relation embeddings parametrized by discrete values, thereby decrease the solution space drastically. Our experiments show that \textit{DihEdral} is able to capture all desired properties such as (skew-) symmetry, inversion and (non-) Abelian composition, and outperforms existing bilinear form based approach and is comparable to or better than deep learning models such as ConvE \cite{convE}.
\end{abstract}

\section{Introduction}
Large-scale knowledge graph (KG) plays a critical role in the downstream tasks such as semantic search \cite{semantic_search}, dialogue management \cite{dialogue} and question answering \cite{kbqa}. In most cases, despite of its large scale, KG is not complete due to the difficulty to enumerate all facts in the real world. The capability of predicting the missing links based on existing dataset is one of the most important research topics for years.  A common representation of KG is a set of triples (\textit{head}, \textit{relation}, \textit{tail}), and the problem of link prediction can be viewed as predicting new triples from the existing set. A popular approach is KG embeddings, which maps both entities and relations in the KG to a vector space such that the scoring function of entities and relations for ground truth distinguishes from false facts \cite{ntn, transe, distmult}. Another family of approaches explicitly models the reasoning process on KG by synthesizing information from paths \cite{guu}. More recently, researchers are applying deep learning methods to KG embeddings so that non-linear interaction between entities and relations are enabled \cite{rgcn, convE}.

The standard task for link prediction is to answer queries (\textit{h}, \textit{r}, ?) or (? \textit{r}, \textit{t}). In this context, recent works on KG embedding focusing on bilinear form methods \cite{complex, hole, analogy, simple} are known to perform reasonably well. The success of this pack of models resides in the fact they are able to model relation (skew-) symmetries. Furthermore, when serving for downstream tasks such as learning first-order logic rule and reasoning over the KG, the learned relation representation is expected to discover relation composition by itself. One key property of relation composition is that in many cases it can be non-commutative. For example, exchanging the order between \verb|parent_of| and \verb|spouse_of| will result in completely different relation (\verb|parent_of| as opposed to \verb|parent_in_law_of|). We argue that, in order to learn relation composition within the link prediction task, this non-commutative property should be explicitly modeled. 

In this paper, we proposed \textit{DihEdral} to model the relation in KG with the representation of dihedral group. The elements in a dihedral group are constructed by rotation and reflection operations over a 2D symmetric polygon.
As the matrix representations of dihedral group can be symmetric or skew-symmetric, and the  multiplication of the group elements can be Abelian or non-Abelian, it is a good candidate to model the relations with all the corresponding properties desired.

To the best of our knowledge, this is the first attempt to employ finite non-Abelian group in KG embedding to account for relation compositions. Besides, another merit of using dihedral group is that even the parameters are quantized or even binarized, the performance in link prediction tasks can be improved over state-of-the-arts methods in bilinear form due to the implicit regularization imposed by quantization.



The rest of paper is organized as follows: in (\S\ref{sec:preliminary}) we present the mathematical framework of bilinear form modeling for link prediction task, followed by an introduction to group theory and dihedral group. In (\S\ref{sec: dihedral}) we formalize a novel model \textit{DihEdral} to represent relations with fully expressiveness. In (\S\ref{sec:training}, \S\ref{sec: experiment}) we develop two efficient ways to parametrize DihEdral and reveal that both approaches outperform existing bilinear form methods.
In (\S\ref{sec: case_studies}) we carried out extensive case studies to demonstrate the enhanced interpretability of relation embedding space by showing that the desired properties of (skew-) symmetry, inversion and relation composition are coherent with the relation embeddings learned from DihEdral.







\section{Preliminaries}\label{sec:preliminary}


\subsection{Bilinear From for KB Link Prediction} \label{subsec:bilinear}

Let $\mathcal{E}$ and $\mathcal{R}$ be the set of entities and relations. A triple $(h, r, t)$, where $\{h, t\}\in \mathcal{E}$ are the head and tail entities, and $r \in \mathcal{R}$ is a relation corresponding to an edge in the KG. 

In a bilinear form, the entities $h$, $t$ are represented by vectors $\bm{h}, \bm{t}\in\mathbb{R}^M$ where $M \in \mathbb{Z}^{+}$, and relation $r$ is represented by a matrix $\bm{R}\in\mathbb{R}^{M\times M}$. The score for the triple is defined as $\phi(h, r, t) = \bm{h}^\top {\bm{R}} \bm{t}$. A good representation of the entities and relations are learned such that the scores are high for positive triples and low for negative triples.



\subsection{Group and Dihedral Group}

Let $g_i, g_j$ be two elements in a set $\mathcal{G}$, and $\odot$ be a binary operation  between any two elements  in $\mathcal{G}$ . The set $\mathcal{G}$ forms a group when the following axioms are satisfied:
\begin{description}[leftmargin=0cm]
\item[Closure] For any two element $g_i, g_j \in \mathcal{G}$, $g_k = g_i \odot g_j$ is also an element in $\mathcal{G}$.
\item[Associativity] For any $g_i, g_j, g_k \in \mathcal{G}$, $(g_i \odot g_j) \odot g_k = g_i \odot (g_j \odot g_k)$.
\item[Identity] There exists an identity element $e$ in $\mathcal{G}$ such that, for every element $g$ in $\mathcal{G}$, the equation $e \odot g = g \odot e = g$ holds.
\item[Inverse] For each element $g$, there is its inverse element $g^{-1}$ such that $g \odot g^{-1} = g^{-1} \odot g = e$.
\end{description}
If the number of group elements is finite, the group is called a \textit{finite group}. If the group operation is commutative, i.e. $g_i \odot g_j = g_j \odot g_i$ for all $g_i$ and $g_j$, the group is called \textit{Abelian}; otherwise the group is \textit{non-Abelian}. 


Moreover, if the group elements can be represented by a matrix, with group operations defined as matrix multiplications, the identity element is represented by the identity matrix and the inverse element is represented as matrix inverse. In the following, we will not distinguish between group element and its corresponding matrix representation when no confusion exists.

A dihedral group is a finite group that supports symmetric operations of a regular polygon in two dimensional space. Here the symmetric operations refer to the operator preserving the polygon. For a $K$-side ($K\in\mathbb{Z}^{+}$) polygon, the corresponding dihedral group is denoted as $\mathbb{D}_{K}$ that consists of $2K$ elements, within which there are $K$ \emph{rotation} operators and $K$ \emph{reflection} operators. A rotation operator $\bm{O}_{k}$ rotates the polygon anti-clockwise around the center by a degree of ${(2\pi m / K)}$, and a reflection operator $\bm{F}_{k}$ mirrors the rotation $\bm{O}_{k}$ vertically.

\begin{figure}[ht]
\begin{centering}
\includegraphics[width=1\columnwidth]{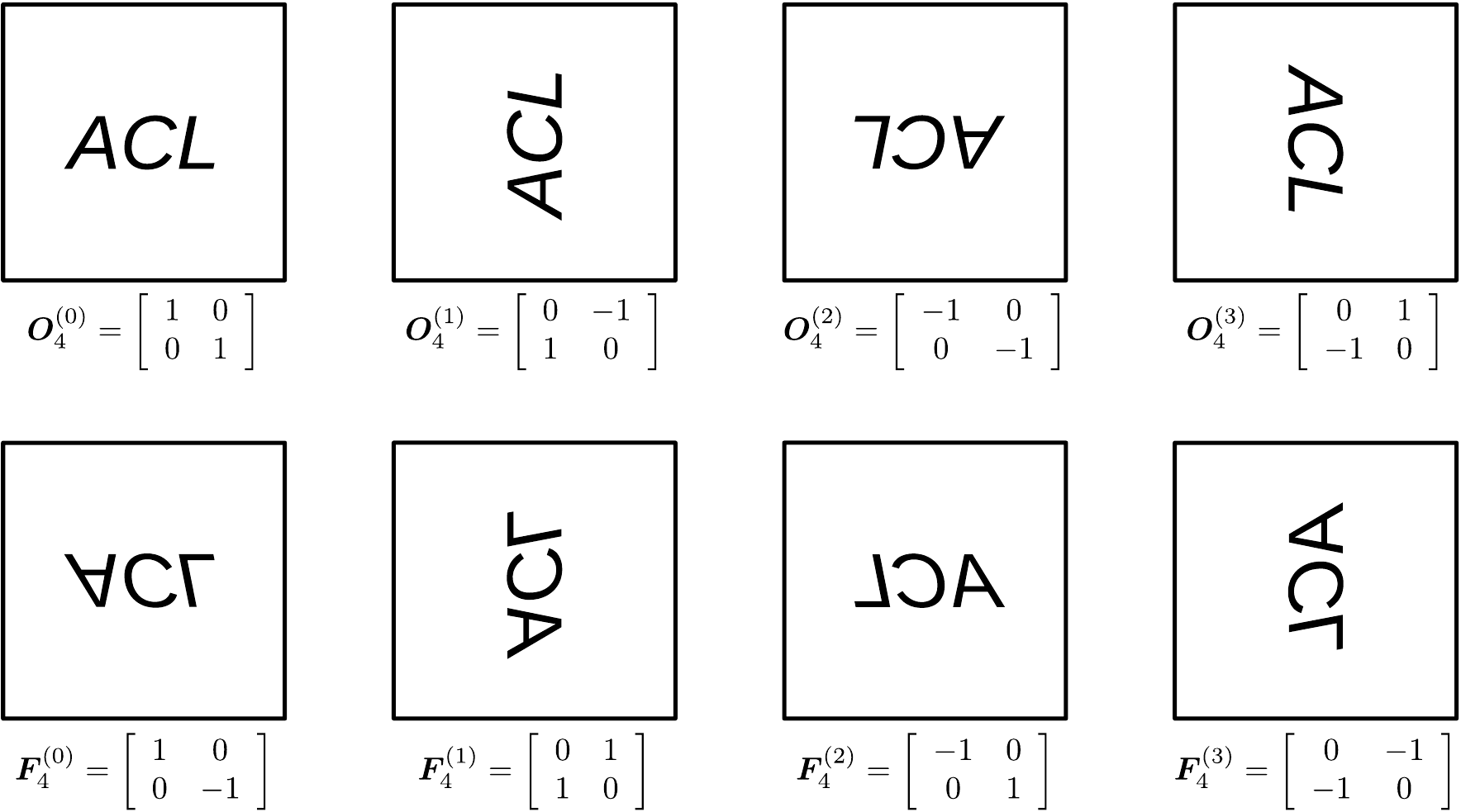}
\par\end{centering}
\caption{Elements in $\mathbb{D}_{4}$. Each subplot represents result after applying the corresponding operator to the square of `ACL' on the upper left corner (on top of $\bm{O}_4^{(0)}$). The top row corresponds to the rotation operators and the bottom row corresponds to the reflection operators.} \label{figure: dihedral}
\end{figure}

The element in the dihedral group $\mathbb{D}_{K}$ can be represented as 2D orthogonal matrices\footnote{There are more than one 2D representations for the dihedral group $\mathbb{D}_{K}$, and we use the orthogonal representation throughout the paper. Check \citealt{group_rep} for details.}:
\begin{equation} \label{eq: D}
\begin{aligned}
\bm{O}_{K}^{(m)} &= \left[\begin{array}{cc}
\cos{\left(\frac{2\pi m}{K}\right)} & -\sin{\left(\frac{2\pi m}{K}\right)}\\
\sin{\left(\frac{2\pi m}{K}\right)} & \cos{\left(\frac{2\pi m}{K}\right)}
\end{array}\right]\\
\bm{F}_{K}^{(m)} &= \left[\begin{array}{cc}
\cos{\left(\frac{2\pi m}{K}\right)} & \sin{\left(\frac{2\pi m}{K}\right)}\\
\sin{\left(\frac{2\pi m}{K}\right)} & -\cos{\left(\frac{2\pi m}{K}\right)}
\end{array}\right]
\end{aligned}
\end{equation}
where $m \in \{0,1,\cdots, K\}$. Correspondingly, the group operation of dihedral group can be represented as multiplication of the representation matrices. Note that when $K$ is evenly divided by $4$, rotation matrices $\bm{O}_K^{(K/4)}$ and $\bm{O}_K^{(3K/4)}$ are skew-symmetric, and all the reflection matrices $\bm{F}_K^{(m)}$ and rotation matrices $\bm{O}_K^{(0)}$, $\bm{O}_K^{(K/2)}$ are symmetric. The representation of $\mathbb{D}_{4}$ is shown in Figure \ref{figure: dihedral}.

\section{Relation Modeling with Dihedral Group and Expressiveness}\label{sec: dihedral}

\begin{table*}[ht]
\centering
\small
\begin{tabular}{l|ccccc}
\bottomrule
\multirow{2}{*}{} & \multirow{2}{*}{Component} & \multirow{2}{*}{Symmetric} & \multirow{2}{*}{Skew-Symmetric} &  \multicolumn{2}{c}{Composition} \\ \cline{5-6} & & & & Abelian & Non-Abelian \\ \hline
\\[-1em]
DistMult& $r_i \in \mathbb{R} $ & \checkmark & $?^{*}$ & \checkmark & NA$^{\dagger}$ \\
\\[-1em]
ComplEx & {$\left[ \begin{array}{cc}
a_{i} & -b_{i}\\
b_{i} & a_{i}
\end{array}\right] $} & {$b_{i}=0$} & {$a_{i}=0$} & \checkmark & NA$^{\dagger}$ \\
\\[-.75em]
ANALOGY  & {
$\left[\begin{array}{cc}
a_i & -b_{i}\\
b_{i} & a_i
\end{array}\right] \cup  
\left\{ c_j \right\}$
} & {$b_{i}=0$} & $a_i, c_j = 0$ & \checkmark & NA$^{\dagger}$ \\
\\[-0.75em]
SimplE & $\left[ \begin{array}{cc}
0 & a_{i}\\
b_{i} & 0
\end{array}\right]$ & $a_{i}=b_{i}$ & $a_{i}=-b_{i}$ & \multicolumn{2}{c}{NA$^{\dagger}$} \\
\\[-1em]
DihEdral & $\mathbb{D}_K$ 
& $
\bm{F}_{K}^{(m)} \cup
\bm{O}_{K}^{(0, K/2)}
$
& $\bm{O}_{K}^{(K/4, 3K/4)} $ & both in $\bm{O}_{K}^{(m)}$ & either in $\bm{F}_{K}^{(m)}$\\
\toprule
\end{tabular}
\caption{Comparison on expressiveness for bilinear KB models. `NA' stands for `not available', and `\checkmark' stands for `always'. ${}^{*}$ DistMult has no skew-symmetric relation representations but it performs well in benchmark datasets because the entity type of head and tails are different. ${}^{\dagger}$ The contents in column `Composition' are subject to the assumption that relation composition corresponds the multiplication of the relation representation. We are not certain if there are other composition rules with which these properties are satisfied.} \label{table:dihedral_property}
\end{table*}



We propose to model the relations by the group elements in $\mathbb{D}_{K}$.
Like ComplEx \cite{complex}, we assume an even number of latent dimensions $2L$. More specifically, the relation matrix takes a block diagonal
form $\bm{R}={\rm diag}\left[\bm{R}^{(1)},\bm{R}^{(2)},\cdots,\bm{R}^{(L)}\right]$
where $\bm{R}^{(l)}\in\mathbb{D}_{K}$ for $l\in\{1,2,\cdots,L\}$.
The corresponding embedding vectors $\bm{h} \in \mathbb{R}^{2L}$ and $\bm{t} \in \mathbb{R}^{2L}$ take the form of $\left[\bm{h}^{(1)}, \cdots, \bm{h}^{(L)}\right]$ and $\left[\bm{t}^{(1)}, \cdots, \bm{t}^{(L)}\right]$ where $\bm{h}^{(l)}, \bm{t}^{(l)} \in \mathbb{R}^2$ respectively. As a result, the score for a triple $(h,r,t)$ in bilinear form can be written as a sum of these $L$ components $\bm{h}^{\top}\bm{R}\bm{t} =\sum_{l=1}^{L}\bm{h}^{(l)\top}\bm{R}^{(l)}\bm{t}^{(l)}$,
We name the model  \textit{DihEdral} because each component $\bm{R}^{(l)}$ is a representation matrix of a dihedral group element. 

\begin{lma}\label{th:orth}
The relation matrix $\bm{R}$ of DihEdral is orthogonal, i.e. $\bm{R}\bm{R}^{\top}=\bm{R}^{\top}\bm{R}=\bm{I}$.
\end{lma}

\begin{lma}\label{th:equiv}
The score of $(h,r,t)$ satisfies $\bm{h}^\top \bm{R} \bm{t} = -\frac{1}{2}\left(\left\Vert\bm{R}^\top\bm{h}-\bm{t}\right\Vert_2^2-\bm{h}^\top \bm{h} -\bm{t}^\top \bm{t}\right)$, consequently maximizing score w.r.t. $\bm{R}$ is equivalent to minimizing $\left\Vert\bm{R}^\top\bm{h}-\bm{t}\right\Vert_2^2$.
\end{lma}

\begin{thm}\label{th:group}
The relations matrices in DihEdral form a group under matrix multiplication.
\end{thm}

Though its relation embedding takes discrete values, DihEdral is fully expressive as it is able to model relations with desired properties for each component $\bm{R}_l$ by the corresponding matrices in $\mathbb{D}_K$.  The properties are summarized in Table \ref{table:dihedral_property}, with comparison to DistMult \cite{distmult}, ComplEx \cite{complex}, ANALOGY \cite{analogy} and SimplE \cite{simple}. \footnote{Note that the condition listed in the table is sufficient but not necessary for the desired property.} The details of expressiveness are described as follows. For notation convenience, we denote $\mathcal{T}^{+} $ all the possible true triples, and $\mathcal{T}^{-}$ all the possible false triples.

\begin{description}[leftmargin=0cm]
\item[Symmetric]  A relation $r$ is symmetric iff $(h, r, t) \in \mathcal{T}^{+} \Leftrightarrow (t,r,h) \in \mathcal{T}^{+}$. Symmetric relations in the real world include \verb|synonym|, \verb|similar_to|.

Note that with DihEdral, the component $\bm{R}_l$ can be a reflection matrix which is symmetric and off-diagonal. This is in contrast to DistMult and ComplEx where the relation matrix has to be diagonal when it is symmetric at the same time.

\item[Skew-Symmetric]  A relation $r$ is skew-symmetric iff $(h, r, t) \in \mathcal{T}^{+}\Leftrightarrow (t,r,h) \in \mathcal{T}^{-}$. Skew-symmetric relations in the real world include \verb|father_of|, \verb|member_of|.

When $K$ is a multiple of $4$, pure skew-symmetric matrices in $\mathbb{D}_4$ can be chosen. As a result, the relation is guaranteed to be skew-symmetric satisfying $\phi(h,r,t)= -\phi(t,r,h)$.


\end{description}
\begin{description}[leftmargin=0cm]
\item[Inversion]  $r_2$ is the inverse of $r_1$ iff $(h, r_1, t) \in \mathcal{T}^{+} \Leftrightarrow (t, r_2, h) \in \mathcal{T}^{+}$. As a real world example, \verb|parent_of| is the inversion of \verb|child_of|.

The inverse of the relation $r$ is represented by $\bm{R}^{-1}$ in an ideal situation: For two positive triples $(h, r_1, t)$ and $(t, r_2, h)$, we have $\bm{R}_1^{\top}\bm{h}\approx \bm{t}$ and $\bm{R}_2^\top\bm{t}\approx \bm{h}$ in an ideal situation  (cf. Lemma \ref{th:equiv}), With enough occurrences of pair $\{h,t\}$ we have $\bm{R}_2=\bm{R}_1^{-1}$.

\item[Composition]  $r_3$ is composition of $r_1$ and $r_2$, denoted as $r_3=r_1\odot r_2$ iff $(h, r_1, m) \in \mathcal{T}^{+} \wedge (m, r_2, t) \in \mathcal{T}^{+} \Leftrightarrow (h, r_3, t) \in \mathcal{T}^{+}$. Example of composition in the real world includes \verb|nationality| = \verb|born_in_city| $\odot$ \verb|city_belong_to_nation|. Depending on the commutative property, there are two cases of relation compositions:
\begin{itemize}
    \item \textbf{Abelian} $r_1$ and $r_2$ are Abelian if $(h, r_1 \odot r_2, t) \in \mathcal{T}^{+} \Leftrightarrow  (h, r_2 \odot r_1, t) \in \mathcal{T}^{+}$. Real world example includes \verb|opposite_gender| $\odot$ \verb|profession| $=$ \verb|profession| $\odot$ \verb|opposite_gender|.
    \item \textbf{Non-Abelian} $r_1$ and $r_2$ are non-Abelian if $(h, r_1 \odot r_2, t) \in \mathcal{T}^{+} \nLeftrightarrow  (h, r_2 \odot r_1, t) \in \mathcal{T}^{+}$. Real world example include \verb|parent_of| $\odot$ \verb|spouse_of| $\ne$ \verb|spouse_of| $\odot$ \verb|parent_of|.
\end{itemize}
\end{description}
In DihEdral, the relation composition operator $\odot$ corresponds to the matrix multiplication of the corresponding representations, i.e. $\bm{R}_3\approx\bm{R}_1\bm{R}_2$.
Consider three positive triples $(h,r_1,m)$, $(m, r_2, t)$ and $(h, r_3, t)$.  In the ideal situation, we have $\bm{R}_1^\top\bm{h}\approx \bm{m}$, $\bm{R}_2^\top\bm{m}\approx \bm{t}$, $\bm{R}_3^\top\bm{h}\approx \bm{t}$ (cf. Lemma \ref{th:equiv}), and further $\bm{R}_2^\top\bm{R}_1^\top\bm{h}\approx\bm{t}$. With enough occurrences of such $\{h,t\}$ pairs in the training dataset, we have $\bm{R}_3\approx\bm{R}_1\bm{R}_2$.

Note that although all the rotation matrices form a subgroup to dihedral group, and hence algebraically closed, the rotation subgroup could not model non-Abelian relations. To model non-Abelian relation compositions at least one reflection matrix should be involved.

\section{Training} \label{sec:training}




In the standard traing framework for KG embedding models,  parameters $\Theta = \Theta_{\mathcal{E}}\cup \Theta_{\mathcal{R}}$, i.e. the union of entity and relation embeddings, are learnt by stochastic optimization methods. For each minibatch of positive triples, a small number of negative triples are sampled by corrupting head or tail for each positive triple, then related parameters in the model are updated by minimizing the binary negative log-likelihood such that positive triples will get higher scores than negative triples.
Specifically, the loss function is written as follows,
\begin{equation} \label{eq: loss}
\min_{\Theta} \sum_{(h,r,t)\in\mathcal{T}^+\cup\mathcal{T}^-}-\log\sigma\left(y\phi(h,r,t)\right)
+ \lambda ||\Theta_{\mathcal{E}}||^2,
\end{equation}
where $\lambda \in \mathbb{R}$ is the $L_2$ regularization coefficient for entity embeddings only, $\mathcal{T}^+$ and $\mathcal{T}^-$ are the sets of positive and sampled negative  triples in a minibatch, and $y$ equals to $1$ if $(h,r,t)\in\mathcal{T}^+$ otherwise $-1$. $\sigma$ is a sigmoid function defined as $\sigma(x)=1/(1+\exp(-x))$.


Special treatments of  the relation representations $\bm{R}$ are required as they takes discrete values. In the next subsections we describe a reparametrization method for general $K$, followed by a simple approach when $K$ takes small integers values. With these treatments, DihEdral could be trained within the standard framework.

\subsection{Gumbel-Softmax Approach}

Each relation component $\bm{R}^{(l)}$ can be parametrized with a one-hot variable $\bm{c}^{(l)}\in \{0,1\}^{2K}$ encoding $2K$ choices of matrices in $\mathbb{D}_K$:
$    \bm{R}^{(l)} = \sum_{k=1}^{2K} c^{(l)}_{k}\bm{D}_k$
where $\{\bm{D}_k, k\in\{1,\cdots,2K\}\}$ enumerates $\mathbb{D}_K$. The number of parameters for each relation is $2LK$ in this approach.


One-hot variable $\bm{c}^{(l)}$ is further parametrized by  $\bm{s}^{(l)} \in \mathbb{R}^{2K}$ by Gumbel trick \cite{gumbel-softmax} with the following steps: 1) take i.i.d. samples $q_1, q_2, \dots, q_{2K}$ from a Gumbel distribution: $q_i = -\log(-\log u_i)$, where $u_i \sim \mathcal{U}(0, 1)$ are samples from a uniform distribution; 2) use log-softmax form of $\bm{s}^{(l)}$ to parametrize $\bm{c}^{(l)}\in \{0,1\}^{2K}$:
\begin{equation} \label{eq: gumbel}
    c^{(l)}_k = \frac{\exp\left[(s^{(l)}_k + q_k) / \tau\right]}{\sum_{k=1}^{2K} \exp\left[(s^{(l)}_k + q_k) / \tau\right]}
\end{equation}
where $\tau$ is the tunable temperature. During training, we start with high temperature, e.g. $\tau_0=3$, to drive the system out of pool local minimums, and gradually cool the system with $\tau=\max(0.5, \tau_0\exp(-0.001 t))$ where $t$ is the number of epochs elapsed. 



\subsection{Reparametrization with Binary Variables}
 Another parametrization technique for $\mathbb{D}_K$ where $K \in \{4, 6\}$ is to parametrize each element in the matrix $\bm{R}^{(l)}$ directly. Specifically we have 
\begin{equation*}
    \bm{R}^{(l)} = \left[
    \begin{array}{cc}
    \lambda & -\alpha \gamma \\
    \gamma & \alpha \lambda \\
    \end{array}\right],
\end{equation*}
where $\lambda = \cos(2\pi k /K)$, $\gamma = \sin(2\pi k /K)$, $k\in \{0,1,\cdots,2K-1\}$ and $\alpha \in \{-1, 1\}$ is the reflection indicator . Both $\lambda$ and $\gamma$ can be parametrized by the same set of binary variables $\{x, y, z\}$:
\begin{align*}
\lambda	& =\begin{cases}
    (x+y)/2 & K=4 \\
    y(3-x)/4 & K=6
\end{cases}, \\
\gamma & =\begin{cases}
    (x-y)/2 & K=4 \\
    z(x+1)\sqrt{3} /4 & K=6
\end{cases}.
\end{align*}

In the forward pass, each binary variable $b\in\{x,y,z\}$ is parametrized by taking a element-wise sign function of a real number: $b = \mathrm{sign}(b_{\mathrm{real}})$ where $b_{\mathrm{real}} \in \mathbb{R}$. 

In the backward pass,  since the original gradient of sign function is almost zero everywhere such that $b_{\mathrm{real}}$ will not be activated, the gradient of  loss with respect to the real variable is estimated with the straight-through estimator (STE)  \cite{ste}. The functional form for STE is not unique and worth profound theoretical study. In our experiments, we used identity STE \cite{hinton_STE}:
\begin{equation*}
    \frac{\partial \mathrm{loss}}{\partial b_{\mathrm{real}}} = \frac{\partial \mathrm{loss}}{\partial b} \mathds{1},
\end{equation*}
where $\mathds{1}$ stands for element-wise identity. 

For these two approaches, we name the model as D$K$-Gumbel for Gumbel-Softmax approach and D$K$-STE for reparametrization using binary variable approach.

\section{Experimental Result} \label{sec: experiment}
This section presents our experiments and results.
We first introduce the benchmark datasets used in our experiments, after that we evaluate our approach in the link prediction task. 

\subsection{Datasets} \label{subsec:datasets}


Introduced in \citet{transe}, WN18 and FB15K are popular benchmarks for link prediction tasks. WN18 is a subset of the famous WordNet database that describes relations between words. In WN18 the most frequent types of relations form reversible pairs (e.g., \verb|hypernym| to \verb|hyponym|, \verb|part_of| to \verb|has_part|). FB15K is a subsampling of Freebase limited to 15k entities, introduced in \citet{transe}. It contains triples with different characteristics (e.g., one to-one relations such as \verb|capital_of| to many-to-many such as \verb|actor_in_film|). YAGO3-10 \cite{convE} is a subset of YAGO3 \cite{yago} with each entity contains at least 10 relations. 


As noted in \citet{fb237, convE}, in the original WN18 and FB15k datasets there are a large amount of test triples appear as reciprocal form of the training samples, due to the reversible relation pairs. Therefore, these authors eliminated the inverse relations and constructed corresponding subsets: WN18RR with 11 relations and FB15K-237 with 237 relations, both of which are free from test data leak. All datasets statistics are shown in Table \ref{tab:data_stats}.

\begin{table}[h]
    \centering
    \small
        \begin{tabularx}{0.44\textwidth}{l|rrrrr} 
        \hline
        Dataset & $|\mathcal{E}|$ & $|\mathcal{R}|$ & Train & Valid & Test \\
        \hline
        WN18 & 41k & 18 &  141k & 5k  &  5k \\ 
        WN18RR & 41k & 11 & 87k  &  3k  &  3k \\
        FB15K & 15k & 1.3k &  483k  &  50k & 59k \\ 
        FB15K-237 & 15k & 237 & 273k & 18k & 20k \\ 
        YAGO3-10 & 123k & 37 & 1M  &  5k  &  5k \\ 
        \hline
        \end{tabularx}
    \caption{Statistics of Datasets.}
    \label{tab:data_stats}
\end{table}

\subsection{Evaluation Metric}
We use the popular metrics \emph{filtered} HITS@{1, 3, 10} and mean reciprocal rank (MRR) as our evaluation metrics as in \citet{transe}.

\begin{table*}[ht]
\centering
\begin{tabular}{lcccc|cccc}
\bottomrule
\multicolumn{1}{l}{\multirow{3}{*}{}} & \multicolumn{4}{c|}{WN18} & \multicolumn{4}{c}{FB15K} \\ \cline{2-9} 
\multicolumn{1}{l}{} & \multicolumn{3}{c}{HITS@N} & \multicolumn{1}{l|}{\multirow{2}{*}{MRR}} & \multicolumn{3}{c}{HITS@N} & \multirow{2}{*}{MRR} \\ \cline{2-4} \cline{6-8}
\multicolumn{1}{l}{} & 1 & 3 & \multicolumn{1}{l}{10} & \multicolumn{1}{l|}{} & 1 & 3 & \multicolumn{1}{l}{10} &  \\ 
\hline
TransE$^{\dagger}$ \cite{transe} & 8.9 & 82.3 & 93.4 & 45.4 & 23.1 & 47.2 & 64.1 & 22.1\\
\hline
DistMult$^{\dagger}$ \cite{distmult} & 72.8 & 91.4 & 93.6 & 82.2 & 54.6 & 73.3 & 82.4 & 65.4\\ 
ComplEx$^{\dagger}$ \cite{complex} & 93.6 & 94.5 & 94.7 & 94.1 & 59.9 & 75.9 & 84.0 & 69.2\\
HolE \cite{hole} & 93.0 & 94.5 & 94.7 & 93.8 & 40.2 & 61.3 & 73.9 & 52.4\\
ANALOGY \cite{analogy} & 93.9 & 94.4 & 94.7 & 94.2 & 64.6 & 78.5 & 85.4 & 72.5\\
Single DistMult \cite{kadlec} & --- & --- & 94.6 & 79.7 & --- & --- & \textbf{89.3} & \textbf{
79.8} \\
SimplE \cite{simple} & 93.9 & 94.4 & 94.7 & 94.2 & \textbf{66.0} & 77.3 & 83.8 & 72.7\\
\hline
R-GCN \cite{rgcn} & 69.7 & 92.9 & \textbf{96.4} & 81.9 & 60.1 & 76.0 & 84.2 & 69.6\\
ConvE \cite{convE} & 93.5 & 94.6 & 95.6 & 94.3 & 55.8 & 72.3 & 83.1 & 65.7\\
\hline
D4-STE & \textbf{94.2} & 94.8 & 95.2 & \textbf{94.6} & 64.1 & \textbf{80.3} & {87.7} & {73.3}\\
D4-Gumbel& \textbf{94.2} & \textbf{94.9} & 95.4 & \textbf{94.6} & 64.8 & 78.2 & 86.4 & 72.8\\

\toprule
\end{tabular}
\caption{Link prediction results on WN18 and FB15K datasets. Results marked by `$\dagger$' are taken from \cite{complex}, and the rest of the results are taken from original literatures.} \label{table: wn18_fb15k}
\end{table*}

\subsection{Model Selection and Hyper-parameters} 
We implemented DihEdral in PyTorch \cite{pytorch}. In all our experiments, we selected the hyperparameters of our model in a grid search setting for the best MRR in the validation set. We trained DK-Gumbel for $K\in\{4, 6, 8\}$ and DK-STE for $K \in \{4, 6\}$ with AdaGrad optimizer \cite{adagrad}, and we didn't notice significant difference in terms of the evaluation metrics when varying $K$. In the following we only report the result for $K=4$.

For D4-Gumbel, we performed grid search for the $L_2$ regularization coefficient $\lambda$ $\in [10^{-5},10^{-4},10^{-3}]$ and learning rate $\in [0.5, 1]$. For D4-STE, hyperparamter ranges for the grid search were as follows: $\lambda$ $\in$ [0.001, 0.01, 0.1, 0.2], learning rate $\in$ [0.01, 0.02, 0.03, 0.05, 0.1]. For both settings we performed grid search with batch sizes $\in$ [512, 1024, 2048] and negative sample ratio $\in$ [1, 6, 10]. We used embedding dimension $2L=1500$ for FB15K, $2L=600$ for both FB15K-237 and YAGO3-10, $2L=200$ for WN18 and WN18RR. We used the standard train/valid/test splits provided with these datasets.



\begin{table*}[h]
\begin{tabular}{lcccc|cccc|cccc}
\bottomrule
\multicolumn{1}{l}{\multirow{3}{*}{}} & \multicolumn{4}{c|}{WN18RR} & \multicolumn{4}{c|}{FB15K-237} & \multicolumn{4}{c}{YAGO3-10} \\ 
\cline{2-13} 
\multicolumn{1}{l}{} & \multicolumn{3}{c}{HITS@N} & \multicolumn{1}{l|}{\multirow{2}{*}{MRR}} & \multicolumn{3}{c}{HITS@N} & \multicolumn{1}{l|}{\multirow{2}{*}{MRR}} & \multicolumn{3}{c}{HITS@N} & \multirow{2}{*}{MRR} \\ 
\cline{2-4} \cline{6-8} \cline{10-12}
\multicolumn{1}{l}{} & 1 & 3 & 10 & \multicolumn{1}{l|}{} & 1 & 3 & 10 & \multicolumn{1}{l|}{} & 1 & 3 & 10 &  \\
\hline
DistMult$^{\dagger}$ & 39.0 & 44.0 & 49.0 & 43.0 & 15.5 & 26.3 & 41.9 & 24.1 & 24.0 & 38.0 & 54.0 & 34.0\\
ComplEx$^{\dagger}$ & 41.0 & 46.0 & 51.0 & 44.0 & 15.8 & 27.5 & 42.8 & 24.7 & 26.0 & 40.0 & 55.0 & 36.0\\
\hline
R-GCN & --- & --- & --- & --- & 15.1 & 26.4 & 41.7 & 24.8 & --- & --- & --- & --- \\
ConvE$^{\dagger}$ & 40.0 & 44.0 & 52.0 & 43.0 & \textbf{23.7} & \textbf{35.6} & 50.1 & \textbf{32.5} & 35.0 & 49.0 & 62.0 & 44.0\\
MINERVA$^{*}$ & 41.3 & 45.6 & 51.3 & 44.8 & 21.7 & 32.9 & 45.6 & 29.3 & --- & --- & --- & --- \\
\hline
D4-STE & \textbf{45.2} & {49.1} & {53.6} & {48.0} & 23.0 & 35.3 & \textbf{50.2} & 32.0 & \textbf{38.1} & \textbf{52.3} & \textbf{64.3}  & \textbf{47.2} \\
D4-Gumbel & 44.2 & \textbf{50.5} & \textbf{55.7} & \textbf{48.6} & 20.4 & 33.2 & 49.6 & 30.0 & 29.4 & 43.6 & 57.3 & 38.8 \\
\toprule
\end{tabular}
\caption{Link prediction results on WN18RR and FB15K-237 datasets. Results marked by `$\dagger$' are taken from \cite{convE}, and result marked by `$*$' is taken from \cite{minerva}.} \label{table: wn18rr_fb15k237_yago}
\end{table*}

 The results of link predictions are shown in Table \ref{table: wn18_fb15k} and \ref{table: wn18rr_fb15k237_yago}, where the results for the baselines are directly taken from original literature. DihEdral outperforms almost all models in bilinear form, and even ConvE in FB15K, WN18RR and YAGO3-10. The result demonstrates that even DihEdral takes discretized value in relation representations, proper modeling the underlying structure of relations using $\mathbb{D}_K$ is essential.

\section{Case Studies} \label{sec: case_studies}
The learned representation from DihEdral is not only able to reach the state-of-the-art performance in link prediction tasks, but also provides insights with its special properties. In this section, we present the detailed case studies on these properties. In order to achieve better resolutions, we increased the embedding dimension to $2L=600$ for WN18 datasets.

\begin{figure}[h]
\includegraphics[width=\columnwidth]{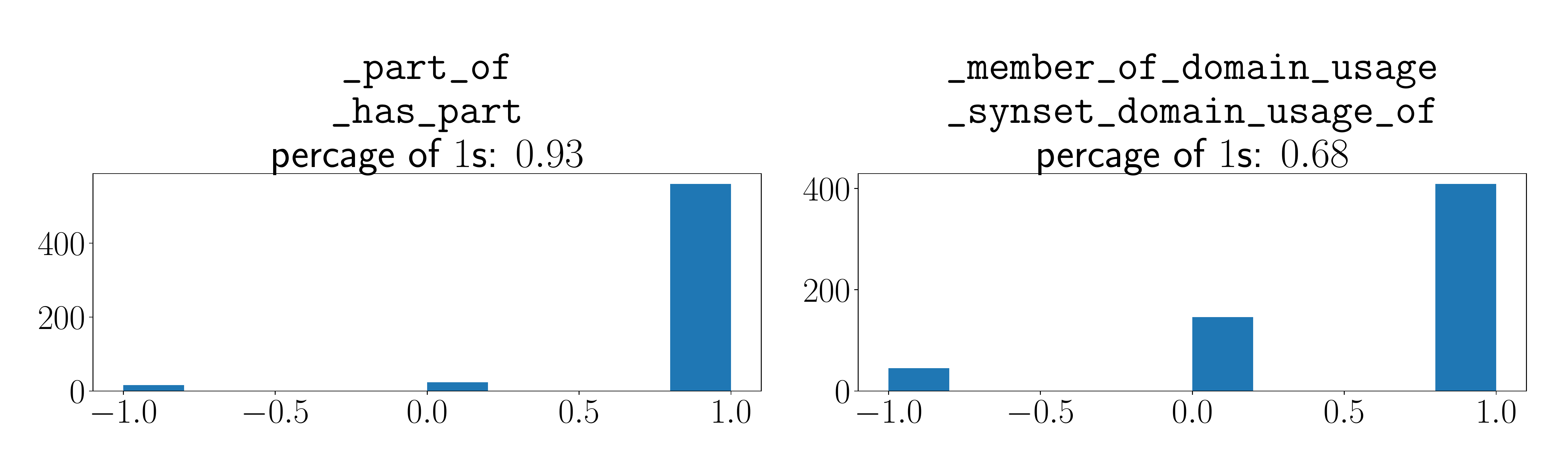}
\includegraphics[width=\columnwidth]{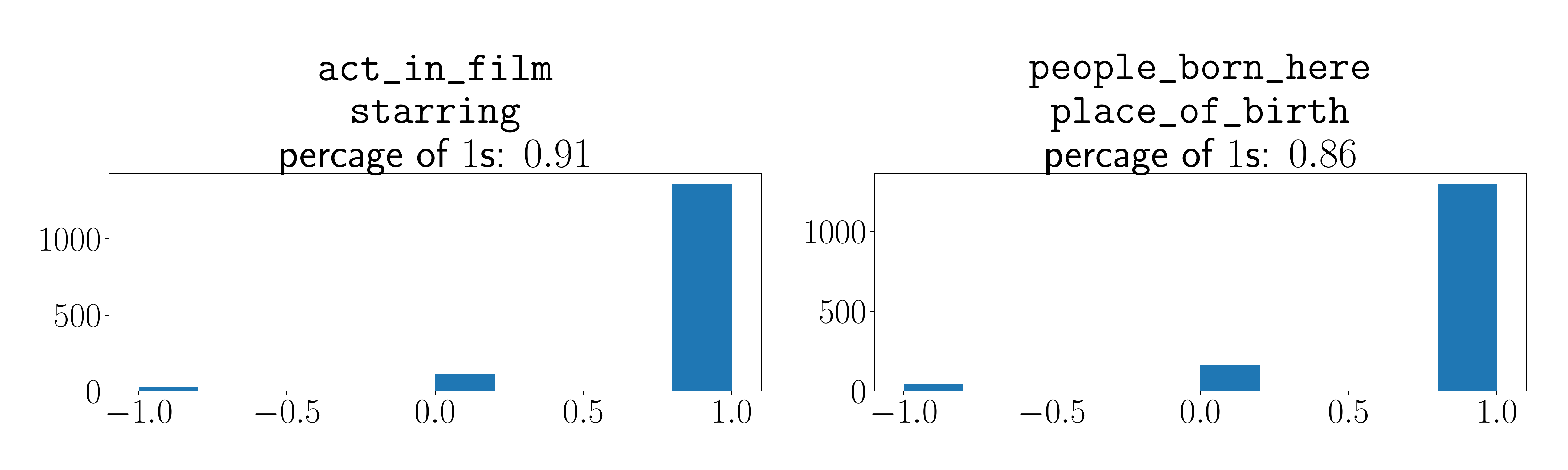}
\caption{Relation inversion in WN18 (top) and FB15K (bottom). Each subplot shows the histogram of diagonal elements in $\bm{R}_1\bm{R}_2$ where $r_1$ is inverse relation of $r_2$. The name of the two relations and the percentage of the $1$s in the diagonal are shown in the first, second and third row of the subplot title, respectively.}\label{fig: inversion}
\end{figure}

\subsection{Inversion} \label{subsec: inversion}
 We show the multiplication of some pairs of inversion relations on WN18 and FB15K in Figure \ref{fig: inversion}, and the result is close to an identity matrix. 
 For the relation pair \{\verb|_member_of_domain_usage|, \verb|_synset_domain_usage_of|\}, the multiplication deviates from ideal identity matrix as the performance for these two relations are poorer compared to the others. 
 We also repeat the same case study for other bilinear embedding methods, however their multiplications are not identity, but close to diagonal matrices with different elements.

\begin{figure}[h]
 \includegraphics[width=1\columnwidth]{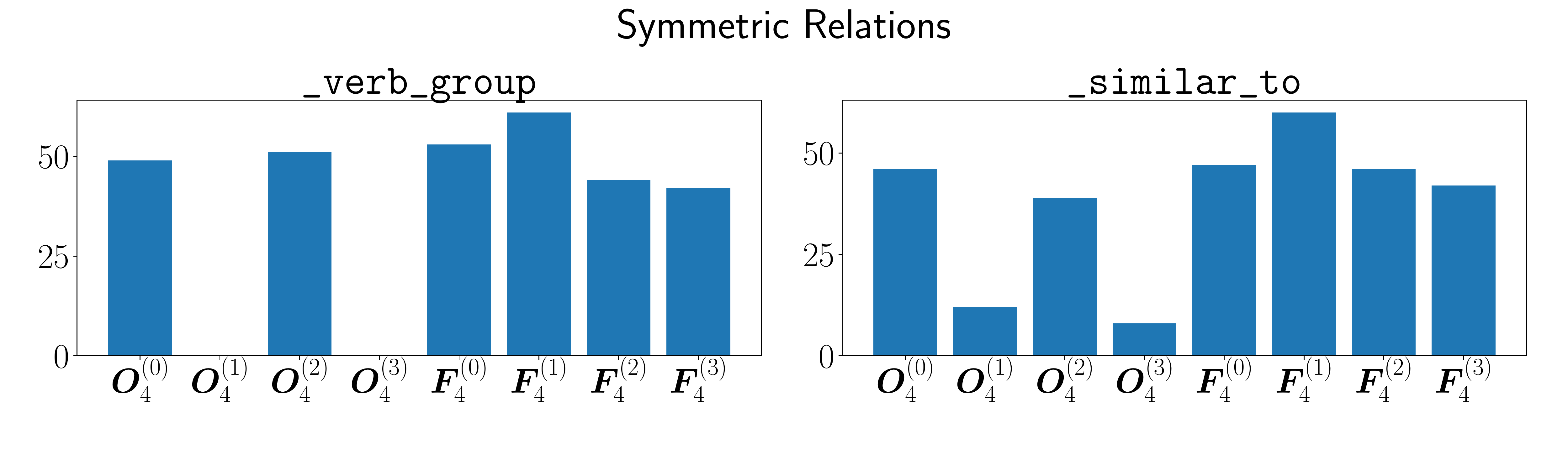}
 \includegraphics[width=1\columnwidth]{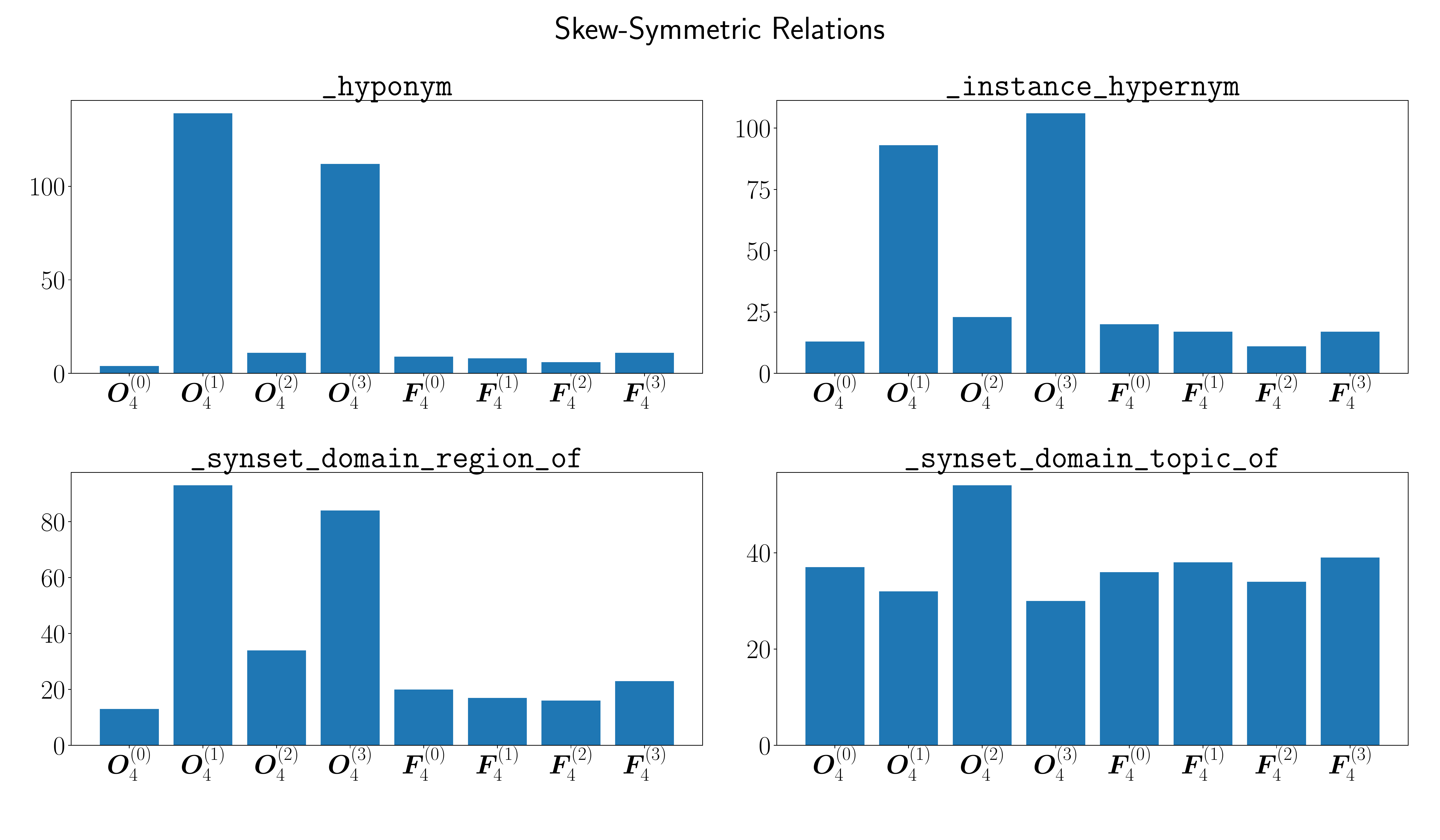}
\caption{Historgram of each component of $\mathbb{D}_4$ for WN18. The top and bottom row corresponds to symmetric and skew-symmetric  relations, respectively. Note that $\bm{O}_4^{(1, 3)}$ are skew-symmetric components and others are symmetric.}\label{figure: sym_skew_sym}
\end{figure}

\subsection{Symmetry and Skew-Symmetry} \label{subsec: skew-symmetry}
Since the KB datasets do not contain negative triples explicitly, there is no penalty to model skew-symmetric relations with symmetric matrices. This is perhaps the reason why DistMult performs well on FB15K dataset in which a lot of relations are skew-symmetric.

To resolve this ambiguity, for each positive triple $(h, r, t)$ with a definite skew-symmetric relation $r$, a negative triple $(t, r, h)$ is sampled with probability 0.5. After adding this new negative sampling scheme in D4-Gumbel, the symmetric and skew-symmetric relations can be distinguished on WN18 dataset without reducing performance on link prediction tasks. Figure \ref{figure: sym_skew_sym} shows that both symmetric and skew-symmetric relations favor corresponding components in $\mathbb{D}_4$ as expected. Again, due to imperfect performance of \verb|_synset_domain_topic_of|, its corresponding representation is imperfect as well. 
We also conduct the same experiment without adding this sampling scheme, the histogram for the symmetric relations are similar, but there is no strong preference for skew-symmetric relations.
\begin{figure}[h]
 \includegraphics[width=1\columnwidth]{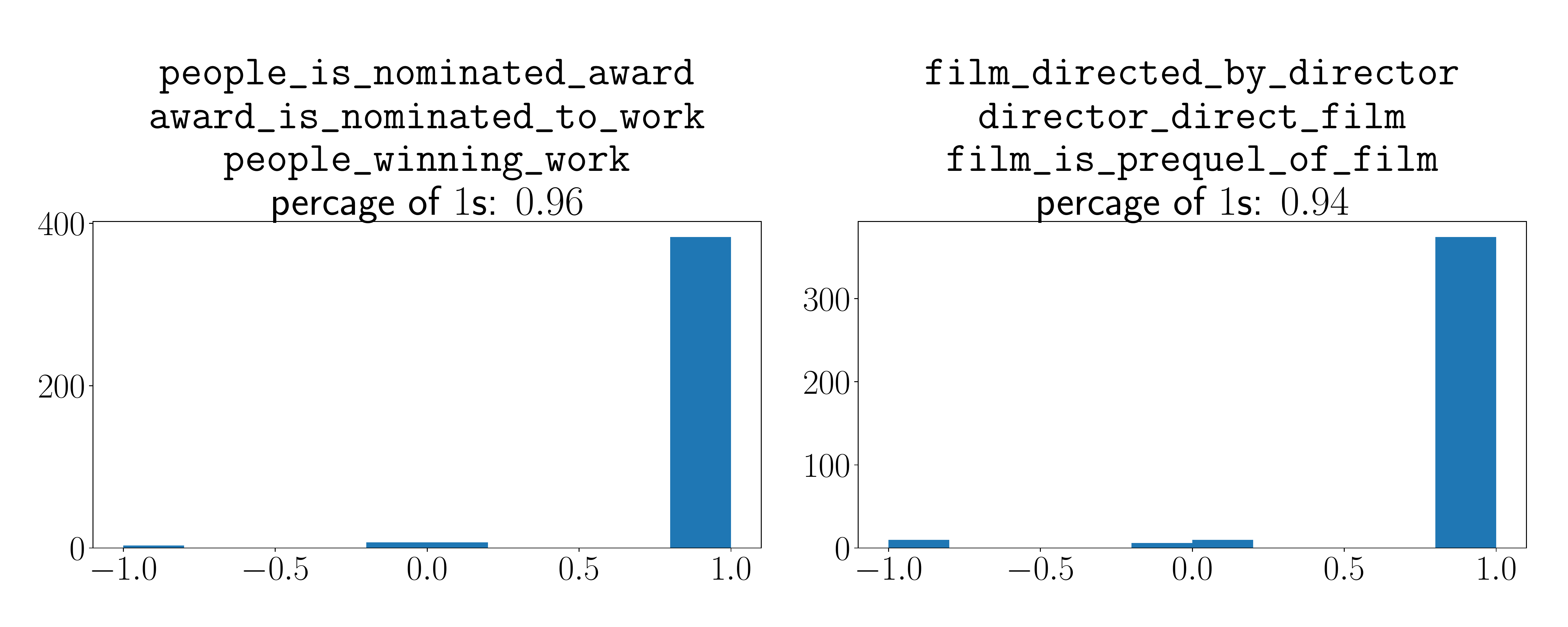}
 \includegraphics[width=1\columnwidth]{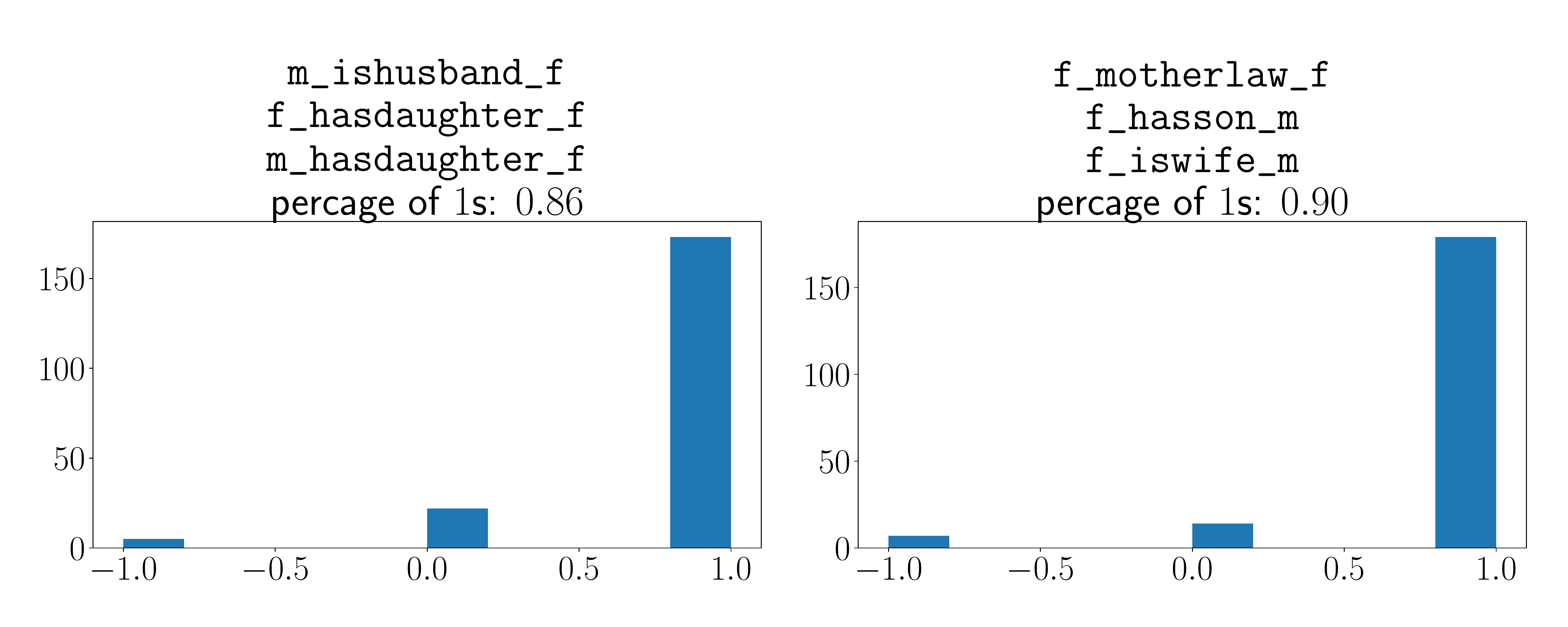}
 \includegraphics[width=1\columnwidth]{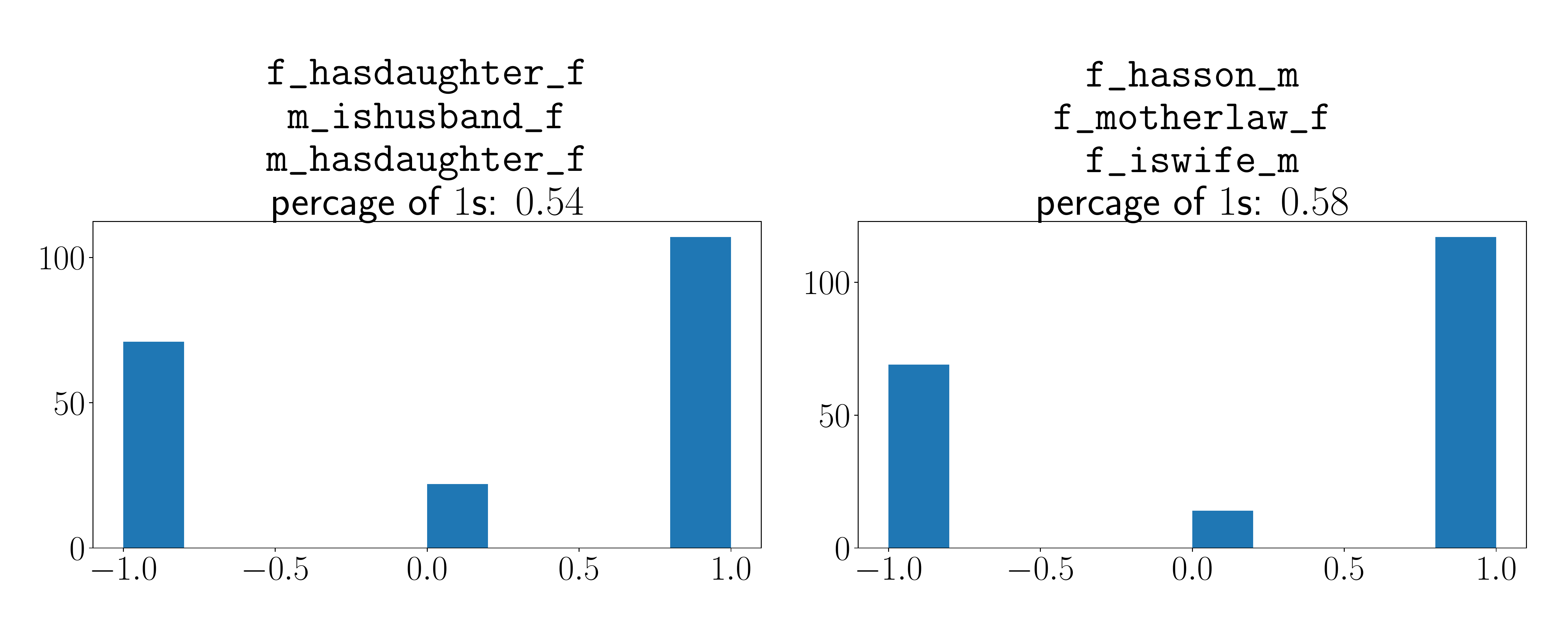}
\caption{Relation composition on FB15K-237 and FAMILY. Each subplot shows the histogram of diagonal elements in $\bm{R}_1\bm{R}_2\bm{R}_3^{-1}$ where $r_3$ is treated as the composition of $r_1$ and $r_2$. The name of the three relations and the percentage of the $1$s in the diagonal are shown in the 1st, 2nd, 3rd and 4th line of subplot title. The two subplots in the first rows shows composition for FB15K-237, and subplots on the second and third row are used to check composition and non-Abelian on FAMILY.}\label{fig: composition}
\end{figure}

\subsection{Relation Composition} \label{subsec: composition} 
In FB15K-237 dataset the majority of patterns is relation composition. However, these compositions are Abelian only because all the inverse relations are filtered out on purpose. To justify if non-Abelian relation compositions can be discovered by DihEdral in an ideal situation, we generate a synthetic dataset called FAMILY. Specifically, we first generated two generations of people with equal number of male and females in each generation, and randomly assigned \verb|spouse| edges within each generation and \verb|child| and \verb|parent| edges between the two generations, after which the \verb|sibling|, \verb|parent_in_law| and \verb|child_in_law| edges are connected based on commonsense logic.

We trained D4-Gumbel on FAMILY with latent dimension $2L=400$. In addition to the loss in Eq. \ref{eq: loss}, we add the following regularization term to encourage the score of positive triple to be higher than that of negative triple for each component independently.
\begin{equation*}
-\sum_{l=1}^{L} \log\sigma\left(\bm{h}^{(l)\top}\bm{R}^{(l)}\bm{t}^{(l)} -\bm{h}^{*(l)\top}\bm{R}^{(l)}\bm{t}^{*(l)}\right).
\end{equation*}
where $(h,r,t)\in \mathcal{T}^{+}$,  and the corresponding negative triple $(h^*,r,t^*)\in \mathcal{T}^{-}$.

For each composition $r_3=r_1\odot r_2$, we compute the histogram of  $\bm{R}_1\bm{R}_2\bm{R}_3^{-1}$ . The result for relation compositions in FB15K-237 and FAMILY is shown in Figure \ref{fig: composition}, from which we could see good composition as matrix multiplication. We also reveal the non-Abelian property in FAMILY by exchanging the order of $r_1$ and $r_2$.
\section{Related Works}

In this section we discuss the related works and their connections to our approach.

TransE \cite{transe} takes relations as a translating operator between head and tail entities. More complicated distance functions \cite{transh, transr, ptranse} are also proposed as extensions to TransE. TorusE \cite{torusE} proposed a novel distance function defined over a torus by transform the vector space by an Abelian group onto a $n$-dimensional torus. ProjE \cite{proje} designs a neural network with a combination layer and a projection layer. R-GCN \cite{rgcn} employs convolution  over multiple entities to capture spectrum of the knowledge graph. ConvE \cite{convE} performs 2D convolution on the concatenation of entity and relation embeddings, thus by nature introduces non-linearity to enhance expressiveness. 

In RESCAL \cite{rescal} each relation is represented by a full-rank matrix. As a downside, there is a huge number of parameters in RESCAL making the model prone to overfitting.  A totally symmetric DistMult \cite{distmult} model simplifies RESCAL by representing each relation with a diagonal matrix. To parametrize skew-symmetric relations, ComplEx \cite{complex} extends DistMult by using complex-valued instead of real-valued vectors for entities and relations. The representation matrix of ComplEx supports both symmetric and skew-symmetric relations while being closed under matrix multiplication. HolE \cite{hole} models the skew-symmetry with circular correlation between entity embeddings, thus ensures shifts in covariance  between embeddings at different dimensions. It was recently showed that HolE is isomophic to ComplEx \cite{isomophic}. ANALOGY \cite{analogy} and SimplE \cite{simple} both reformulate the tensor decomposition approach in light of analogical and reversible relations.

Though embedding based approach achieves state-of-the-art performance on link prediction task, symbolic relation composition is not explicitly modeled. In contrast, the latter goal is currently popularized by directly modeling the reasoning paths \cite{composition_vectors, deeppath, minerva, multi_hop, longterm}. As paths are consistent with reasoning logic structure, non-Abelian composition is supported by nature. 

DihEdral is more expressive when compared to other bilinear form based embedding methods such as DistMult, ComplEX and ANALOGY. As the relation matrix is restricted to be orthogonal, DihEdral could bridge translation based and bilinear form based approaches as the training objective \textit{w.r.t.} the relation matrix is similar (cf Lemma \ref{th:equiv}). Besides, DihEdral is the first embedding method to incorporate non-Abelian relation compositions in terms of matrix multiplications (cf. Theorem \ref{th:group}).

\section{Conclusion}


This paper proposed \textit{DihEdral} for KG relation embedding. By leveraging the desired properties of dihedral group,  relation (skew-) symmetry, inversion, and (non-) Abelian compositions are all supported. Our experimental results on benchmark KGs showed that DihEdral outperforms existing bilinear form models and even deep learning methods.  Finally, we demonstrated that the above g properties can be learned from DihEdral by extensive case studies, yielding a substantial increase in interpretability from existing models. 

\section*{Acknowledgments}
The authors would like to thank Vivian Tian, Hua Yang, Steven Li and Xiaoyuan Wu for their supports, and anonymous reviewers for their helpful comments.

\bibliography{dihedral}
\bibliographystyle{acl}
\end{document}